%% file: main.tex
\pdfoutput=1

\documentclass[11pt]{article}

\usepackage[]{acl}

\usepackage{times}
\usepackage{latexsym}

\usepackage[T1]{fontenc}

\usepackage[utf8]{inputenc}

\usepackage{microtype}

\usepackage{inconsolata}

\title{Decoupled Vocabulary Learning Enables Zero-Shot Translation from Unseen Languages}

\author{Carlos Mullov\textsuperscript{1} \and Ngoc-Quan Pham\textsuperscript{1} \and Alexander Waibel\textsuperscript{1,2} \\
  \textsuperscript{1}Karlsruhe Institute of Technology, Karlsruhe, Germany \\
  \textsuperscript{2}Carnegie Mellon University, Pittsburgh PA, USA \\
  \texttt{\{firstname\}.\{lastname\}@kit.edu}}

\usepackage{amsmath}
\usepackage{amssymb}
\usepackage{amsthm}
\usepackage{booktabs}
\usepackage{listings} 

\usepackage{rotating}
\usepackage{subcaption}
\usepackage{multicol}
\usepackage{multirow}
\usepackage{array}
\usepackage{hyperref}
\usepackage[inline]{enumitem}
\usepackage{colortbl}
\usepackage{changepage}
\usepackage{tikz}
\usetikzlibrary{tikzmark,calc}

\DeclareMathAlphabet{\mathcal}{OMS}{cmsy}{m}{n}

\newcommand{\lang}{\ell}
\newcommand{\new}{\text{new}}
\newcommand{\newlang}{\lang_{\new}}

\newcommand{\baselangs}{\mathcal{L}_{\text{base}}}
\newcommand{\dmodel}{d_{\text{model}}}

\newcommand{\grcll}{\cellcolor{gray!30}}

\begin{document}
\maketitle
\input{front/abstract.tex}%
\input{main/introduction.tex}
\input{main/background.tex}
\input{main/related_work.tex}
\input{main/approach.tex}
\input{main/evaluation.tex}
\input{main/conclusion_future_work.tex}
\input{back/limitations.tex}
\input{back/acknowledgements.tex}

\bibliography{anthology,custom}

\clearpage
\appendix
\input{appendix/appendix.tex}

\end{document}

%% file: front/abstract.tex

\begin{abstract}
    Multilingual neural machine translation systems learn to map sentences of different languages into a common representation space.
    Intuitively, with a growing number of seen languages the encoder sentence representation grows more flexible and easily adaptable to new languages.
    In this work, we test this hypothesis by zero-shot translating from unseen languages.
    To deal with unknown vocabularies from unknown languages we propose a setup where we decouple learning of vocabulary and syntax, i.e. for each language we learn word representations in a separate step (using cross-lingual word embeddings), and then train to translate while keeping those word representations frozen.
    We demonstrate that this setup enables zero-shot translation from entirely unseen languages. 
    Zero-shot translating with a model trained on Germanic and Romance languages we achieve scores of 42.6 BLEU for Portuguese-English and 20.7 BLEU for Russian-English on TED domain.
    We explore how this zero-shot translation capability develops with varying number of languages seen by the encoder.
    Lastly, we explore the effectiveness of our decoupled learning strategy for unsupervised machine translation.
    By exploiting our model's zero-shot translation capability for iterative back-translation we attain near parity with a supervised setting.
\end{abstract}

%% file: main/introduction.tex

\section{Introduction}

In order to extend Machine Translation from a hundred languages to the 7,000 languages of the world new methods have to be developed with more economical use of data from individual languages.
This requires better use of abstractions and representations of languages to facilitate transfer learning to low-resource languages.
In machine translation, it had been attempted by developing linguistic structures explicitly in classical Interlingua systems \citep{levin1998interlingua} or imposing structure on statistical models \citep{suhm1994towards,wang1998modeling}.
However, developing such structures was labor intensive and did not achieve acceptable performance or required domain restrictions.
First attempts toward learning latent representations of language \citep{wang1991connectionist,wang1995connectionist} still required linguistic structures or specific design.
Successful performance on open-domain translation was only feasible later using multilingually trained end-to-end neural translation models \citep{bahdanau14,johnson-etal-2017-googles,ha2016}.
These models share hidden representations between languages and allow for multilingual translation that benefits low-resource languages by transfer learning from high-resource languages and enable zero-shot translation on new language directions.

The multilingual encoder in such models will ideally learn to map semantically similar sentences onto similar latent representations -- even across different languages.
Ongoing research in the field of zero-shot translation shows that multilingual NMT models exhibit this property to a certain extent \citep{Duquenne:2023:sonar_arxiv} and that enforcing the similarity between sentence representations across different languages in turn also improves the zero-shot capabilities \citep{pham-etal-2019-improving,arivazhagan2019,zhang-etal-2020-improving,liu-etal-2021-improving-zero,pan-etal-2021-contrastive}.
At the same time, cross-lingual transfer learning research shows that massively multilingual systems can be rapidly extended to new languages on very little data \citep{neubig-hu-2018-rapid,artetxe-etal-2020-cross,garcia-etal-2021-towards,marchisio-etal-2023-mini,chen2023improving}.
This suggests that exposing the NMT model to a higher number of languages increases the plasticity of its sentence representation and enables easier adaptation to unseen languages.

When extending a multilingual model by an unseen language, however, naturally the question arises of how to deal with the unknown vocabulary words.
The field of incremental learning devised several strategies for adapting to new vocabularies.
\citet{artetxe-etal-2020-cross} show that simply relearning the embedding layer can be done efficiently, compared to training the model from scratch.
\citet{garcia-etal-2021-towards} find that retraining the shared BPE codes once the new language data become available allows one to keep most of the pre-trained vocabulary, making even more efficient adaptation possible.

In this work, we go one step further and show that by designing a system where word representations are learned in a separate step we can zero-shot translate from an unknown language without any adaptation whatsoever.
We therefore employ cross-lingual word embeddings \citep{artetxe-etal-2016-learning,lample2018word} -- a technique that has extensively been used in unsupervised machine translation, but hasn't so far been given much consideration for incremental learning -- to align monolingual word representations into a common space, and then use these as our regular word embedding layer in the NMT system.
Crucially, keeping these embeddings frozen in the NMT training allows us to align new language vocabularies into the common embedding space post-training.

Furthermore, we show that this zero-shot translation capability allows us to easily generate synthetic parallel data for unknown languages and enables simple and efficient unsupervised machine translation without the need for denoising autoencoding on massive amounts of monolingual data.

Finally, we also experiment on continuous output NMT \citep{kumar2018vmf}, which -- while underperforming the softmax models in terms of BLEU scores -- perfectly fits our setup and attains vastly superior performance in terms of training times.

We make our code and training configurations available under \url{https://github.com/cmullovisl/clwe-transfer} and we release our pre-trained multilingual models and embedding alignments.

%% file: main/background.tex

\section{Multilingual Neural Machine Translation}%
\label{sec:machine_translation}
In training the multilingual NMT system we aim to estimate the probability $\mathbb{P}(Y_{\lang_{tgt}} = y \mid X_{\lang_{src}} = x)$ that the sentence $y$ in the target language $\lang_{tgt}$ is a suitable translation of the source sentence $x$ in the source language $\lang_{src}$.
We describe the distribution of the sentences in different natural languages $\lang$ through random variables $X_{\lang}$, $Y_{\lang}$.
A universal encoder maps the sentences from the different input distributions onto a single, shared latent distribution $H_{\text{enc}} = enc(X_{\lang})$.
A decoder is then tasked to model the probability distribution from this latent variable: $\mathbb{P}(Y_{\lang_{tgt}} = y \mid H_{\text{enc}} = c)$. 
Identically to a regular bilingual NMT system, the multilingual translation system consists of a neural network, which, in a supervised setting, is trained to maximize the likelihood of a dataset of parallel sentences $\mathcal{D} = \{(x_1, y_1), \ldots{}, (x_n, y_n)\}$, where sentence pairs $(x_i, y_i)$ are translations of each other:
\[
    \min_{\vartheta \in \Theta}{
        \left\{-\sum_{i=1}^{n}{
            \log{
                \mathbb{P}\left( X_{\lang_{tgt}^i} = y_i \mid X_{\lang_{src}^i} = x_i, \vartheta \right)
            }
        }\right\}
    }
\]%
Note that -- aside from the added target language selection mechanism -- the only difference to bilingual translation is the nature of the input and output distribution.
Thus, the main practical difference in the training a multilingual system is that instead of using a bilingual training corpus, $\mathcal{D}$ consists of a concatenation of several parallel corpora.

%% file: main/related_work.tex

\section{Related Work}
\paragraph{Cross-Lingual Word Embeddings}
\citet{qi-etal-2018-pre} look into the effectiveness of pre-trained embeddings for regular supervised NMT and find them to be helpful in a multilingual setting.
\citet{kim-etal-2019-effective} conduct unsupervised MT by transferring the parameters learned on a high resource language pair to a new source language.
Similar to our work they therefore align the new language embeddings to the parent model source embeddings, but they focus on a bilingual setting.
Compared to their work we show that this setup not only allows for unsupervised MT but zero-shot translation from unseen languages, especially when combined with a multilingual setup.

\paragraph{Unsupervised MT}
Our work is closely related to unsupervised NMT \citep{lample2018unsupervised,lample-etal-2018-phrase,artetxe2017,artetxe-etal-2019-effective,conneau-etal-2020-unsupervised,song2019mass,liu-etal-2020-multilingual-denoising}.
Initial works train unsupervised NMT systems by bootstrapping a common representation space between two languages through a combination of cross-lingual word embeddings and denoising autoencoding, and then exploit these common representations for iterative back-translation \citep{lample2018unsupervised}.
Through unsupervised means they thus induce a shared sentence representation, essentially bootstrapping a multilingual NMT system from monolingual data.
Later works do away with the cross-lingual word embeddings and create this shared representation space through large scale denoising autoencoding on massive amounts of monolingual data \citep{song2019mass,liu-etal-2020-multilingual-denoising} and multilinguality \citep{sen-etal-2019-multilingual,siddhant-etal-2020-leveraging,conneau-etal-2020-unsupervised,liu-etal-2020-multilingual-denoising}.
In contrast to this development, we show that cross-lingual word embeddings remain valuable when combined with regular NMT training, allowing to bootstrap high-quality multilingual representations even on small-scale datasets.

\paragraph{New Language Learning}
Multiple works look into adding unseen languages to multilingual encoder-decoder systems.
\citet{neubig-hu-2018-rapid} concerned about the possibility of rapidly extending a multilingual NMT model by an unseen low-resource language.
They compare between bilingual training, multilingual training alongside a similar source language and multilingual pre-training with as many languages as possible.
They come to the conclusion that pre-training in a highly multilingual setting -- in their case 58 source languages -- significantly improves transferability to low-resourced languages.
In particular, they achieve significant performance (up to 15.5 BLEU) on an entirely unseen language, provided parallel training data on a closely related language.
In this work we use the same parallel data for NMT training and significantly improve upon their result through the employment of cross-lingual word embeddings.
\citet{escolano-etal-2019-bilingual} devise an approach to multilingual NMT with independent encoders and decoders that enables addition of new languages.
\citet{li-etal-2023-multilingual} fine-tune a model based on XLM-R and use it to zero-shot translate from languages seen in pre-training, but unseen in the supervised downstream task.
\citet{DBLP:journals/corr/abs-2008-00401} look into extending mBART25 \citep{liu-etal-2020-multilingual-denoising} by 25 new languages and show that given enough modelling capacity one can add new languages without loss in quality on existing ones.
In their work, however, they only consider new languages sufficiently represented in the base model vocabulary, and they leave the original vocabulary unmodified.

\paragraph{Vocabulary Adaptation}
\citet{kim-etal-2019-effective}, \citet{garcia-etal-2021-towards}, \citet{marchisio-etal-2023-mini} and \citet{chen2023improving} look into vocabulary adaptation for new languages.
\citet{garcia-etal-2021-towards} show that rebuilding the subword vocabulary once the new language data is available results in most of the vocabulary matching the old one.
Retaining the embedding vectors for common subword units allows for quick adaptation and minimal forgetting of the existing languages.
In our work we directly compare to this method and show that a vocabulary based on aligned pre-trained embeddings not only presents a viable alternative to a subword-based one, but also allows for more efficient adaptation to and zero-shot translation from unseen languages.

%% file: main/approach.tex

\section{Generalizing to Unseen Languages}
To learn a multilingual NMT system that easily generalizes to unseen languages we propose a decoupled learning approach, divided into multiple steps:
\begin{enumerate}
    \setlength{\itemsep}{0pt}
    \item
        \label{enum:setup-step-1}
        train word embeddings on monolingual data for each of the \emph{base languages} $\ell \in \baselangs$
    \item align the word embeddings into a common embedding space 
    \item initialize the encoder embedding layer of a standard NMT model with the learned word representations
    \item supervised NMT training on $\baselangs$ parallel data with frozen encoder word embeddings
\end{enumerate}
In this setup we may later extend the translation model by an unseen language $\newlang \notin \baselangs$ by learning word representations for $\newlang$ and aligning them into the common word embedding space (see Section~\ref{sub:cross_lingual_word_embeddings}).
We show that this alignment alone enables us to start translating from $\newlang$ in a plug-and-play fashion, even without any additional updates to the NMT model's main body (i.e. the Transformer layers).

\begin{figure}[tpb]
    \centering
    \includegraphics[width=0.96\linewidth]{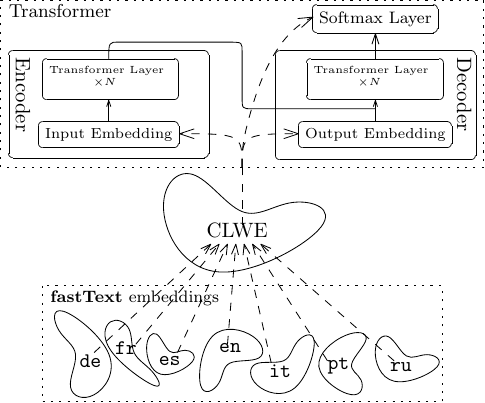}
    \caption{Our NMT architecture consists of a Transformer model with pre-trained cross-lingual word embeddings (CLWE) for embedding layers.
             The embedding vectors are obtained through alignment of monolingual \emph{fastText} embeddings for each language into a common embedding space.}%
    \label{fig:clwe-architecture}
\end{figure}

Optionally, the learned word representations may also be employed on the decoder-side, which will further enable the back-translation-based adaptation detailed in Section~\ref{sub:unsupervised_machine_translation}.
Figure \ref{fig:clwe-architecture} describes our decoupled learning NMT architecture.%

\subsection{Decoupled Vocabulary Learning}
\label{sub:cross_lingual_word_embeddings}
In conventional training of a universal multilingual NMT system the $\newlang$ word vectors are randomly initialized and then trained end-to-end on the NMT objective.
In training the model then learns to represent its words in a shared multilingual embedding space, by learning cross-lingual word correlations from the parallel data.
In a monolingual data only setting we aim to emulate these multilingual word representations through cross-lingual word embeddings.
These word embeddings are then integrated into our encoder-decoder NMT model by simply using them in the word embedding layers. 

For each language $\lang \in \baselangs \cup \{\newlang\}$ we use pre-trained monolingual word embeddings learned with \emph{fastText} \citep{bojanowski-etal-2017-enriching}, which we then align into one common space through cross-lingual embedding alignment \citep{joulin-etal-2018-loss}.
For a common alignment between multiple languages we choose to align into a \emph{single hub space} \citep[SHS][]{heyman-etal-2019-learning}, i.e.\ we pick one of our base languages as a \emph{pivot} and align each of the embedding spaces $E_{\lang}$ to $E_{\text{pivot}}$.

\subsection{Zero-Shot Translation from Unseen Languages}
\label{sub:zero_shot_translation_from_unseen_languages}
\begin{figure}[tpb]
    \centering
    \includegraphics[width=1.0\linewidth]{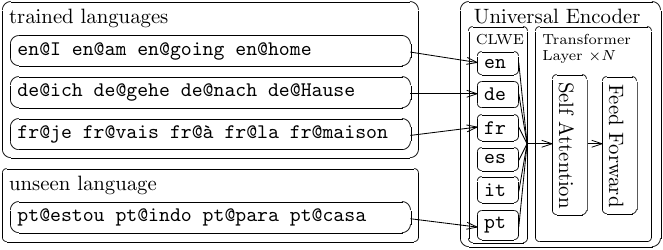}
    \caption{
        Our decoupled learning of word representations enables us to zero-shot translate from an unseen language $\newlang$ in a plug-and-play fashion.
        The Transformer layers have no prior exposure to $\newlang$.
        The cross-lingual word embeddings (CLWE) serve as lookup table for $\newlang$ words.
    }%
    \label{fig:blind_decoding}
\end{figure}
To provide an indication of how well the universal encoder generalizes an unseen language $\newlang$ we let our multilingual model decode from $\newlang$ sentences, without any sorts of prior exposure to $\newlang$.
As outlined in Figure~\ref{fig:blind_decoding} we therefore employ the cross-lingual embeddings to look up the $\newlang$ word vectors and then simply decode from the resulting sequence.
While the encoder may not know the syntax of the unseen language, many syntactical concepts in language -- such as grammatical case, part-of-speech, grammatical genders, or tense -- are encoded at the word level\footnote{Take German for example, a noun would have different surface forms for different grammatical cases, which the fastText model should take into account and appropriately assign different word vectors.}.
As such we hypothesize that the word vectors trained on the new language alone provide enough syntactic level information to perform the language comprehension task to a certain degree.

To test our hypothesis that a high degree of multilinguality will help in making the sentence representations more general we can then vary the number of languages the encoder gets exposed to in training.
We may then observe how this affects languages closely related to languages in $\baselangs$ as well as languages that are distant to any of the languages in $\baselangs$.

\subsection{Unsupervised Machine Translation}%
\label{sub:unsupervised_machine_translation}
Iterative back-translation \citep{sennrich-etal-2016-improving,lample2018unsupervised,artetxe-etal-2018-unsupervised} is a common approach in unsupervised NMT, where a bidirectional model is used to generate synthetic parallel data and then subsequently adapted on this data to then generate even better synthetic data.
To iteratively improve translation between two languages $\ell_1$ and $\ell_2$ one would
\begin{enumerate}
    \setlength{\itemsep}{0pt}
    \item
        \label{enum:iterative_back_translation_step_1}
        translate $\ell_1$ monolingual data to $\ell_2$ to generate $\ell_2\rightarrow\ell_1$ parallel data
    \item adapt the model on the $\ell_2\rightarrow\ell_1$ data
    \item translate $\ell_2$ monolingual data to $\ell_1$ to generate $\ell_1\rightarrow\ell_2$ parallel data
    \item adapt the model on the $\ell_1\rightarrow\ell_2$ data
\end{enumerate}
The resulting model will be able to produce better $\ell_2\rightarrow\ell_1$ synthetic data than in Step~\ref{enum:iterative_back_translation_step_1} and the process may be repeated for iterative improvements until convergence.

To obtain usable translations for Step~\ref{enum:iterative_back_translation_step_1} the model, however, must first be bootstrapped, which is usually done through denoising autoencoding or language modelling on massive amounts of monolingual data.
This process is both costly and oftentimes fails for languages that are either too distant or have too little monolingual data available \citep{kim-etal-2020-unsupervised,marchisio-etal-2020-unsupervised}.
We show that in our setup we may skip this step entirely by zero-shot translating from an unknown language $\newlang$ to kick-start the iterative back-translation process.
Furthermore, we show that through our model's improved ability to generalize to unseen languages we can robustly improve $\newlang$ translation quality, even when $\newlang$ is so distant to any other language $\lang \in \baselangs$ that the zero-shot generated synthetic data is barely intelligible, and even on relatively small amounts of monolingual $\newlang$ data.

%% file: main/evaluation.tex

\section{Experiments}%
\subsection{General Experiment Setup}%
\label{sec:experiment_setup}
\paragraph{Cross-lingual word embeddings}
As the basis for all of our monolingual word embeddings we use pre-trained \emph{fastText} models\footnote{\url{https://fasttext.cc/docs/en/crawl-vectors.html}} trained on Common Crawl \citep{wenzek-etal-2020-ccnet}.
These models provide 300-dimensional high-quality embeddings for full word units, with pre-trained models available in 157 languages.
Using the pivot approach described in section \ref{sub:cross_lingual_word_embeddings} we align these embeddings into a common embedding space using direct supervised optimization on the RCSLS criterion \citep{joulin-etal-2018-loss} using the bilingual dictionaries provided by the MUSE project \citep{lample2018word}.
To reduce the vocabulary size, we subsequently regenerate our vocabulary and the corresponding embedding vectors using only the words in our NMT training corpus.
For alignment accuracies and vocabulary sizes see Appendix Table~\ref{tab:vocab_size}.

\paragraph{Full word-units}%
\label{par:full_word_units}
While using a joint BPE lets us share information between vocabularies and reduces the size of the multilingual vocabulary, for our approach we opt for full word units.
Our preliminary experiments have shown full-word units to outperform our BPE models when used in combination with frozen cross-lingual embeddings.
Furthermore, this gives us better compatibility with distant languages with separate alphabets and avoids the problem of having to extend the learned BPE codes post-training when we add a new language to the model.
To alleviate the OOV word problem we use \emph{fastText} as the basis for our monolingual word embeddings, which, by using subword level information, gives us the ability to map unseen words into the embedding space.

From an implementation perspective we merge our monolingual vocabularies and concatenate their embedding matrices, while dealing with duplicate words across different languages by encoding each word with a language specific prefix, e.g.\ we encode the English word \emph{bank} as \emph{\textbf{en@}bank}.
At inference-time we limit the target vocabulary to the language we decode to, thereby forcing the model to only choose tokens from our desired target language, which is not trivial in a model based on BPE units shared across multiple languages.
Due to the large vocabulary sizes using full word-units increases the number of parameters immensely.
However, since we keep the embedding layers frozen the amount of trainable parameters and GPU memory consumption\footnote{for non-trainable parameters we do not have to save the Adam optimizer states} \emph{decreases} in practice.
Furthermore, our approach results in lower sequence lengths, thus partially making up for higher computational complexity of the larger softmax embedding layer.
At inference-time (and potentially also training time) the computational complexity may further be reduced through dynamic vocabulary sub-selection \citep{senellart-etal-2018-opennmt}.
We leave this line of research for future work.

\paragraph{Von Mises-Fisher loss}%
\label{par:von_mises_fisher_loss}
Since the fastText embeddings we employ in the output vocabulary are normalized to unit norm, the matrix multiplication in the final projection layer is similar to a cosine-similarity search through the fastText embedding space.
Due to our large vocabulary sizes this makes the computation of the output layer during training and test time more costly.
To counter this issue we conduct experiments with von Mises-Fisher loss for NMT \citep{kumar2018vmf}, where model outputs are optimized toward frozen pre-trained word embedding targets, which perfectly fits our setup.
This loss is similar to the cosine-loss and it makes the computational complexity of training entirely independent of the vocabulary size, and we observe up to 5.8$\times$ speedups in training throughput.
We find this approach to consistently under-perform the traditional softmax-based models (Appendix~\ref{appendix:basesystem_scores}), but in light of the speedup we keep this approach for the Section~\ref{sub:effect-of-multilinguality} experiments.
In the following we refer to this model variant as \texttt{vMF}.

\paragraph{Datasets}%
\label{par:datasets}
As our training, development and evaluation data sets we use multilingual transcriptions of TED talks in 60 languages \citep{qi-etal-2018-pre}.
For our main body of experiments we include English, German, Spanish, French and Italian as the supervised languages, which we refer to as \emph{Romance-Germanic} languages.
We train the basesystem on 20 translation directions, and a parallel corpus size of $3,576,046$ sentences.
For our experiments on unsupervised MT we focus on monolingual in-domain data obtained from the same corpus of TED talks.
To prevent contamination this corpus is split into train, validation and test data using the same TED talks across all languages \citep{qi-etal-2018-pre}.
For larger scale out-of-domain experiments we use monolingual data from the Wiki-40B dataset \citep{guo-etal-2020-wiki} and newscrawl data \citep{bojar-etal-2014-findings} and we further evaluate on Flores \citep{goyal-etal-2022-flores} and Tatoeba \citep{tiedemann-2020-tatoeba}.

\paragraph{NMT model}%
\label{evaluation:nmt_model}%
\begin{table*}[!tpb]
    \centering
    \input{tables/blind-translation.tex}
\end{table*}
For our NMT model we use a Transformer model \citep{vaswani2017} with relative position encodings \citep{shaw-etal-2018-self} and post-layernorm.
In accordance with our 300-dimensional fastText embeddings we train 9-layer models with embedding dimension 300 unless otherwise specified.
Our implementation is based on the repository provided by \citet{kumar2018vmf}\footnote{\url{https://github.com/Sachin19/seq2seq-con}}, which is based on \emph{OpenNMT-py} \citep{klein-etal-2017-opennmt}.
We share embedding parameters across the encoder and the decoder wherever applicable, and use tied output embeddings.
Refer to Appendix~\ref{sec:training_parameters} for the full set of training parameters.

We train our softmax-based models using Adam \citep{DBLP:journals/corr/KingmaB14} and our vMF model \citep{kumar2018vmf} using RAdam \citep{DBLP:conf/iclr/LiuJHCLG020} for 160,000 updates or until early stopping.
For estimates of training times and GPU hours see Appendix~\ref{sec:estimate_of_gpu_hours}.
We evaluate our models using the ChrF++ metric \citep{popovic-2017-chrf} and list BLEU \citep{papineni-etal-2002-bleu} scores in the appendix.
The supervised direction evaluation scores for our Romance-Germanic models are listed in Table~\ref{tab:basemodel_comparison}.

Finally, to verify the viability of our setup -- namely a full word-unit translation system with frozen pre-trained word embeddings -- we also train a standard subword-unit translation system which matches our proposed model as closely as possible.
We train the BPE model on the concatenated multilingual corpus for 50k merge operations.
The results show an average score of 55.1/51.5/55.3 ChrF++ for the softmax-based model, the vMF model and the BPE baseline (detailed in Table~\ref{tab:basemodel_comparison}), demonstrating that while the vMF model lags behind in terms of translation performance, the softmax-based model closely matches the baseline despite the word embeddings not being optimized for translation.

To ensure better reproducibility we base our data and training pipelines on \texttt{DVC} \citep{ruslan_kuprieiev_2024_10637615}. To avoid data leakage of unseen languages into supervised training we structure our experiments in two stages, where the choice of $\newlang$ may be postponed until the second stage.

\subsection{Zero-Shot}%
\label{sub:eval_zero_shot_translation}
In the following we present our results for translating from unseen languages after aligning their word embeddings into the common word embedding space and plugging them into the NMT model's encoder without any further adaptation.
We report the results in Table~\ref{tab:blind_translation}.
For comparison, we also evaluate the translation performance of a regular BPE-based model on unseen languages and observe that it fails to produce reasonable translations.
Similar to the findings of \citet{neubig-hu-2018-rapid} we observe up to 12.9 BLEU (34.9 ChrF++) on Portuguese-English, but find the translations to be unintelligible.

For additional comparison we employ \citeauthor{garcia-etal-2021-towards}'s \citeyearpar{garcia-etal-2021-towards} method of adapting the model via vocabulary substitution.
Matching their findings we observe that rebuilding the subword vocabulary with the new language results in 77\,-\,95\,\% overlap with the old vocabulary and after adaptation the translation performance nearly matches the oracle setup.
We further improve on their method by freezing the decoder (but training the cross-attention weights) when adding a new source language, improving by +0.8 ChrF++ (+0.6 BLEU) on Portuguese-English.
Compared to the resulting 62.4 ChrF++ after adaptation we achieve up to 61.0 ChrF++ translating zero-shot on our in-domain test set.
On the out-of-domain test sets we even out-perform the adapted system by +3.1 ChrF++ on Flores (27.5 vs 30.6) and +1.8 ChrF++ on Tatoeba (36.9 vs 38.7). 
Note, that this best performing model is our Romance-Germanic model with several supervised languages close to Portuguese.
For the same model we observe much higher gaps between supervised and zero-shot performance:
For Russian the baseline out-performs our method by 4 ChrF++, which is, however, reasonable considering that our model has at the time of decoding never been exposed to any Slavic language.

With growing distance to the supervised languages this gap grows higher.
For Korean the English translations end up being barely intelligible\footnote{Note, that we have observed some tokenization issues with Korean, leading to large vocabulary size of 330k. Generally, we observe drops in performance when fastText requires special tokenization, i.e. for Chinese, Japanese, Vietnamese.}.
Nevertheless, the translations are still good enough to kick-start the iterative back-translation process, as we show in Section~\ref{sub:unsupervised_machine_translation}.
We observe that on more distant languages the model oftentimes gets stuck in decoding loops, which we address in a post-processing step to remove repeated n-grams.

\begin{table*}[!htpb]
    \centering
    \input{tables/backtranslation.tex}
\end{table*}

\paragraph{Effect of Multilinguality}
\label{sub:effect-of-multilinguality}
To study the effect of multilinguality we train multilingual models on varying numbers of languages and families and compare the translation performance on unseen languages in Table~\ref{tab:blind_translation}.
On the supervised languages we see a clear trend of performance degradation the more languages we include in our model, e.g. for Spanish$\rightarrow$English we see clear degradation when going from bilingual to multilingual.
For the zero-shot directions on the other hand we see a clear boost in performance when going from bilingual to multilingual.
As we increase the number of languages, however, we do not see a clear trend of improvement.
Rather than the number of languages, the relatedness of the added languages plays the more important role, i.e. adding non-Romance to the model continuously degrades performance on Portuguese, and on Slavic languages the performance increases once we include Slavic languages in training and decreases once we include Asiatic languages.

Finally, for the three languages aren't directly related to any of the supervised languages (i.e. Arabic, Turkish, Korean) we see a jump in performance upon inclusion of Slavic languages, but no consistent improvement on inclusion of Asiatic languages.

\subsection{Unsupervised Machine Translation}%
\label{sub:unsupervised_machine_translation}
In Section~\ref{sub:eval_zero_shot_translation} we show that our setup enables translation from unknown languages, albeit with lower quality the further the distance from our training languages.
In our final experiment we exploit this capability for zero-shot iterative back-translation \citep{lample2018unsupervised}.
We therefore simply perform beam search to translate the monolingual target language portion of our TED corpus to each of the known languages to generate synthetic parallel data.
On this synthetic data we then train our softmax-based Romance-Germanic model for 16,000 steps or until saturation.
In training, we freeze the encoder in each odd iteration, the decoder in each even iteration.
Table~\ref{tab:newlang_results} reports the ChrF++ scores after each iteration of training on back-translated data (Table~\ref{tab:newlang_results_bleu} for BLEU scores), alongside the supervised fine-tune on real bilingual data.
We observe that for the closest language (Portuguese) a single round of back-translation closes the gap, and on the out-of-domain test set even outperforms the supervised fine-tune.
On the more distant languages we see a rapid closing of gap between the supervised and the unsupervised fine-tune.

Note that our primary goal is to demonstrate that our setup enables us to skip the expensive bootstrapping process involving denoising autoencoding on massive amounts of monolingual data exhibited in most approaches to unsupervised MT.
As such we don't continue the iterative back-translation process until saturation, but execute at most 6 iteration.
We believe that more iterations will further close the gap between supervised and unsupervised performance.

\paragraph{Effect of Domain}
Many analyses of unsupervised MT have shown the importance of the considered domain.
As the Table~\ref{tab:newlang_results} bottom row shows the monolingual data we use for our in-domain experiments ranges from 52k to 208k sentences, which can be considered extremely small amounts when it comes to monolingual data.
To study the effect of domain and amount of data we perform training on Wikipedia and news domain in Appendix~\ref{sec:effect_of_domain}.
We observe that changing the domain helps in some cases, for improving the overall translation performance a larger number of iterations on TED domain, however, seem to be most effective.

Finally, we compare to the approaches detailed in \citet[MASS][]{song2019mass} and \citet{artetxe-etal-2019-effective}.
For both systems we use their provided implementations to train an English-Turkish unsupervised MT system, using 1947k sentences of Turkish Wikipedia data for bootstrapping.
Both systems end up failing to produce any reasonable translations on any of the test sets.
To verify our setup we further use MASS to train an English-French unsupervised MT model on the French portion of our Wikipedia data.
This model ends up achieving 52.4 (54.1) ChrF++ from (to) English on the Flores test data.

%% file: tables/blind-translation.tex
\begin{adjustwidth}{-0.25cm}{}
\setlength{\tabcolsep}{4pt}
\begin{tabular}{l|rc|rrrrrrrrrrr}
    \toprule
    Trained on         &\small $|\baselangs|$ &\small $\dmodel$ & de-en & es-en & ru-en & vi-en & pt-en & nl-en & bg-en & ar-en & tr-en & ko-en \tabularnewline
    \toprule
    \multicolumn{13}{l}{\textbf{Softmax Models}} \tabularnewline
    \toprule
    \tikzmark{bpe_from}%
    BPE baseline                           & 5 & 300 & \grcll 57.9  & \grcll 62.7  & 11.0  & 12.4  & 34.9  & 23.4  & 11.4  &  9.9  & 16.1  & 11.1 \tabularnewline
    \tikzmark{bpe_to}%
    \citep{garcia-etal-2021-towards}      & 5 & 300 & \grcll & \grcll & \grcll 47.4 & \grcll 48.9 & \grcll 62.4  & \grcll 57.3 & \grcll 59.2 & \grcll 50.9 & \grcll 48.2 & \grcll 41.7 \tabularnewline
    \midrule
    \tikzmark{softmax_from}%
    \small en,de,es,fr,it                  & 5 & 300 & \grcll 57.5  & \grcll 62.6  & 42.4  & 35.1  & 61.2  & 51.8  & 49.9  & 38.8  & 32.5  & 25.1 \tabularnewline
    \small en,de,es,fr,it,ru,uk,pl,vi,zh   &10 & 576 & \grcll 56.1  & \grcll 61.5  & \grcll 48.1  & \grcll 48.2  & 60.4  & 51.2  & 53.2  & 41.7  & 34.8  & 25.0 \tabularnewline
    \midrule
    \tikzmark{softmax_to}%
    supervised fine-tune                   & 5 & 300 & \grcll & \grcll & \grcll 48.5  & \grcll 46.9 & \grcll 63.3  & \grcll 56.5  & \grcll 58.5 & \grcll 52.6 & \grcll 48.2  & \grcll 39.1 \tabularnewline
    \toprule
    \multicolumn{13}{l}{\textbf{vMF Models} \citep{kumar2018vmf}} \tabularnewline
    \midrule
    \texttt{es}$\rightarrow$\texttt{en}    & 1 & 300 & 36.0  & \grcll 60.9  & 35.1  & 27.5  & 54.7  & 40.5  & 33.9  & 31.2  & 26.9  & 21.6 \tabularnewline
    \texttt{es}$\leftrightarrow$\texttt{en}& 2 & 300 & 40.9  & \grcll 59.9  & 37.1  & 30.0  & 55.2  & 42.4  & 42.1  & 34.3  & 29.0  & 22.6 \tabularnewline
    \midrule
    \small en,de,fr,it                     & 4 & 300 & \grcll 54.3  & 55.1  & 39.6  & 30.1  & 56.0  & 47.6  & 45.5  & 33.8  & 29.2  & 21.4 \tabularnewline
    \tikzmark{vmf_from}%
    \small en,de,es,fr,it                  & 5 & 300 & \grcll 54.3  & \grcll 59.5  & 40.0  & 30.8  & 58.1  & 47.9  & 45.7  & 34.9  & 29.4  & 20.4 \tabularnewline
    \small en,de,es,fr,it,ru,uk,pl         & 8 & 300 & \grcll 52.7  & \grcll 57.8  & \grcll 45.2  & 32.8  & 56.4  & 47.9  & 50.5  & 37.1  & 31.2  & 22.3 \tabularnewline
    \small en,de,es,fr,it,ru,uk,pl,vi,zh   &10 & 300 & \grcll 51.9  & \grcll 57.2  & \grcll 44.6  & \grcll 45.4  & 55.2  & 47.5  & 49.4  & 37.0  & 31.0  & 23.0 \tabularnewline
    \midrule
    \tikzmark{vmf_to}%
    supervised fine-tune                   & 5 & 300 & \grcll 54.4  & \grcll 59.9  & \grcll 45.5  & \grcll 44.3  & \grcll 60.4 & \grcll 53.5 & \grcll 56.3 & \grcll 48.4 & \grcll 47.9 & \grcll 33.4 \tabularnewline
    \bottomrule
\end{tabular}
\end{adjustwidth}
\caption{
    ChrF++ scores for decoding from an unseen language on the TED data test split \citep{qi-etal-2018-pre}.
    Grey colored cells indicate supervised training on the source language, white cells indicate no prior exposure to the source language.
    $\dmodel$ denotes the Transformer model dimension and arrows indicate what models are used as basis for fine-tuning.
    Note, that every column in the three fully supervised rows represents a separate bilingual fine-tune. 
    For BLEU scores and translation directions other than English see Appendix~\ref{appendix:supplementary_blind_translation_scores}.
}
\label{tab:blind_translation}

\begin{tikzpicture}[overlay,remember picture]
\draw[->] ($(pic cs:bpe_from)+(-4pt,.5ex)$) to ($(pic cs:bpe_from)+(-10pt,.5ex)$) to ($(pic cs:bpe_to)+(-10pt,+0.5ex)$) to ($(pic cs:bpe_to)+(-4pt,+0.5ex)$);
\draw[->] ($(pic cs:softmax_from)+(-4pt,.5ex)$) to ($(pic cs:softmax_from)+(-10pt,.5ex)$) to ($(pic cs:softmax_to)+(-10pt,+0.5ex)$) to ($(pic cs:softmax_to)+(-4pt,+0.5ex)$);
\draw[->] ($(pic cs:vmf_from)+(-4pt,.5ex)$) to ($(pic cs:vmf_from)+(-10pt,.5ex)$) to ($(pic cs:vmf_to)+(-10pt,+0.5ex)$) to ($(pic cs:vmf_to)+(-4pt,+0.5ex)$);
\end{tikzpicture}

%% file: tables/backtranslation.tex
\setlength{\tabcolsep}{2pt}
\begin{tabular}{l rcl r rcl r rcl r rcl}
    \toprule
    \multirow{2}{*}{Model}             & \multicolumn{3}{c}{en$\rightarrow$pt} & & \multicolumn{3}{c}{en$\rightarrow$ru} & & \multicolumn{3}{c}{en$\rightarrow$tr} & & \multicolumn{3}{c}{en$\rightarrow$ko} \tabularnewline
    \cline{2-4}\cline{6-8}\cline{10-12}\cline{14-16}
                                       & \small TED & \small Flores & \small Tatoeba & & \small TED & \small Flores & \small Tatoeba & & \small TED & \small Flores & \small Tatoeba & & \small TED & \small Flores & \small Tatoeba \tabularnewline
    \midrule
    \citep{song2019mass}  &       &       &       & &        &       &       & &  23.4  & 29.7  & 20.0  & &        &       &      \tabularnewline
    \citep{artetxe-etal-2019-effective}
                          &       &       &       & &        &       &       & &  10.8  & 12.1  & 12.5  & &        &       &      \tabularnewline
    \midrule
    Iteration 1           & 59.6  & 53.3  & 56.8  & &  41.4  & 39.9  & 48.4  & &  35.1  & 34.7  & 35.9  & &  15.1  & 12.8  & 13.6 \tabularnewline
    Iteration 3           & 60.0  & 53.7  & 57.2  & &  42.9  & 41.9  & 50.9  & &  38.7  & 37.7  & 42.4  & &  18.3  & 14.7  & 15.6 \tabularnewline
    Iteration 5           & 60.0  & 53.8  & 57.2  & &  43.3  & 42.5  & 52.9  & &  40.4  & 38.9  & 45.6  & &  19.5  & 15.3  & 16.3 \tabularnewline
    \midrule
    Supervised            & 59.0  & 52.8  & 56.0  & &  42.8  & 41.5  & 51.3  & &  42.2  & 39.8  & 48.6  & &  20.6  & 15.2  & 17.4 \tabularnewline
    \toprule

    \multirow{2}{*}{Model}             & \multicolumn{3}{c}{pt$\rightarrow$en} & & \multicolumn{3}{c}{ru$\rightarrow$en} & & \multicolumn{3}{c}{tr$\rightarrow$en} & & \multicolumn{3}{c}{ko$\rightarrow$en} \tabularnewline
    \cline{2-4}\cline{6-8}\cline{10-12}\cline{14-16}
                                       & \small TED & \small Flores & \small Tatoeba & & \small TED & \small Flores & \small Tatoeba & & \small TED & \small Flores & \small Tatoeba & & \small TED & \small Flores & \small Tatoeba \tabularnewline
    \midrule
    \citep{song2019mass}  &       &       &       & &        &       &       & &  24.5  & 31.0  & 20.5  & &        &       &      \tabularnewline
    \citep{artetxe-etal-2019-effective}
                          &       &       &       & &        &       &       & &   8.7  &  9.3  &  8.1  & &        &       &      \tabularnewline
    \midrule
    Zero-shot             & 61.2  & 56.3  & 60.5  & &  42.4  & 40.9  & 46.0  & &  32.5  & 32.9  & 28.4  & &  25.1  & 25.9  & 21.3 \tabularnewline
    Iteration 2           & 62.6  & 57.4  & 62.4  & &  46.1  & 45.5  & 52.1  & &  40.4  & 40.0  & 41.0  & &  31.8  & 31.8  & 27.4 \tabularnewline
    Iteration 4           & 62.4  & 57.5  & 62.3  & &  47.1  & 47.3  & 55.4  & &  43.0  & 41.7  & 46.7  & &  33.1  & 33.3  & 29.5 \tabularnewline
    Iteration 6           & 62.3  & 57.6  & 62.1  & &  47.1  & 47.4  & 55.8  & &  44.5  & 42.7  & 49.9  & &  33.6  & 33.8  & 30.5 \tabularnewline
    \midrule
    Supervised            & 63.3  & 56.7  & 62.3  & &  48.5  & 46.5  & 57.3  & &  48.2  & 44.2  & 54.4  & &  39.1  & 34.8  & 40.1 \tabularnewline
    \bottomrule
    \# of Sentences        & \multicolumn{3}{c}{52k} & & \multicolumn{3}{c}{208k} & & \multicolumn{3}{c}{182k} & & \multicolumn{3}{c}{206k} \tabularnewline
    \bottomrule
\end{tabular}

\caption{
    ChrF++ scores after each iteration of back-translation.
    In each iteration we use our Romance-Germanic multilingual model to zero-shot translate from TED domain to create synthetic parallel data.
}
\label{tab:newlang_results}

%% file: main/conclusion_future_work.tex

\section{Conclusion}
In this work, we have looked into the generalization ability of a multilingual NMT system on unseen languages.
We therefore propose a decoupled learning setup for MT training, where we learn the word representations of each language in a separate step.
This enables easier generalization to new languages as this approach enables extension of the vocabulary post-training.
We show that through simple alignment of an unseen language word representations into the common vocabulary space, we can produce intelligible translations without any further adaptation.
We further show that on languages similar to the ones the NMT system has already been exposed to, this zero-shot translation setup even approaches supervised translation performances.

Lastly we show that this capability to produce translations for unseen languages can be exploited for efficient unsupervised machine translation.

\paragraph{Future Work}
In the future we would like to explore the efficacy of our presented setup for an incremental learning system, which would entail in-depth experiments on catastrophic forgetting.

%% file: back/limitations.tex

\section{Limitations}
The method described in this work relies on high quality word embeddings and alignments between them.
Training such high-quality word embeddings requires access to large amounts of monolingual data, which may be available in abundance for high and mid-resource languages, but may be unavailable in the amounts necessary for endangered languages, or languages which are primarily spoken, such as dialects or languages without an official orthography.
This issue might be dealt with through data-efficient means of training word embeddings, which might entail an adaptation of word embeddings from a related high-resource language.
Furthermore, we rely on bilingual dictionaries for learning high-quality alignments between embedding spaces.
This may present a limitation for low-resource scenarios, in practice, however, bilingual dictionaries are less of an issue than the availability of monolingual data \citep[see][]{kamholz-etal-2014-panlex,vulic-etal-2019-really}.

In part due to the availability of high-quality multi-parallel data and in part due to a limit in computational resources our work focuses on relatively small scale experiments.
Furthermore, we observe that our models are highly tuned to the TED domain and perform poorly on out-of-domain test sets in the considered supervised, as well as unsupervised settings.

Lastly, our setup operates on word-level vocabularies, which presents a computational bottleneck in the output vocabulary.
This makes the model difficult to train and deploy in its current implementation.
In this work we look into accelerating training using the approach presented by \citet{kumar2018vmf} but we observe a drop in translation quality in doing so.
These issues may, however, be addressed with more engineering work, such as using language specific mini-batches (in combination with gradient accumulation), and a dynamic vocabulary sub-selection at training time and at inference time \citep{senellart-etal-2018-opennmt}.

%% file: back/acknowledgements.tex

\section{Acknowledgements}
This work is supported from the European Union’s Horizon research and innovation programme under grant agreement No 101135798, project Meetween (My Personal AI Mediator for Virtual MEETtings BetWEEN People).
Part of this work was supported by funding from the pilot program Core-Informatics of the Helmholtz Association (HGF).

%% file: appendix/appendix.tex

\section{Vocabulary Details}
\begin{table}[h!tpb]
    \centering
    \begin{tabular}{c|rr}
        \toprule
        $\ell$ & vocab size & CLWE acc. \tabularnewline
        \hline
        en &  75,234 & 1.000 \tabularnewline
        de & 122,287 & 0.554 \tabularnewline
        es &  93,683 & 0.750 \tabularnewline
        fr &  79,633 & 0.732 \tabularnewline
        it &  96,506 & 0.696 \tabularnewline
        \hline
        total & 467,343 &    \tabularnewline
        \hline
        ar & 203,093 \tabularnewline
        bg & 129,866 & 0.628 \tabularnewline
        cs & 116,719 & 0.642 \tabularnewline
        ja &  54,344 \tabularnewline
        ko & 331,292 \tabularnewline
        nl & 102,027 & 0.645 \tabularnewline
        pl & 164,762 \tabularnewline
        pt &  45,483 & 0.734 \tabularnewline
        ro & 117,749 \tabularnewline
        ru & 178,587 & 0.627 \tabularnewline
        tr & 209,492 \tabularnewline
        uk & 121,020 \tabularnewline
        vi &  48,261 \tabularnewline
        zh &  77,637 \tabularnewline
        \bottomrule
    \end{tabular}
    \caption{The number of words in the vocabulary generated from the training corpus for each of the languages.
             \emph{total} represents the size of the shared vocabulary of the base model.
             The column \emph{CLWE acc} describes the nearest neighbour accuracies for the cross-lingual embedding alignments to $\lang_{\text{pivot}} = \mathit{en}$.}
    \label{tab:vocab_size}
\end{table}

\section{Training Parameters}
\label{sec:training_parameters}
    \begin{tabular}{l|r}
        \toprule
        \multicolumn{2}{l}{\textbf{General Settings}} \tabularnewline
        \midrule
        layers                                  & 9 \tabularnewline
        rnn-size                                & 300 \tabularnewline
        word-vec-size                           & 300 \tabularnewline
        transformer-ff                          & 1200 \tabularnewline
        heads                                   & 6 \tabularnewline
        warmup-init-lr                          & 1e-8 \tabularnewline
        warmup-end-lr                           & 0.0007 \tabularnewline
        min-lr                                  & 1e-9 \tabularnewline
        encoder-type                            & transformer \tabularnewline
        decoder-type                            & transformer \tabularnewline
        param-init-glorot                       & True \tabularnewline
        label-smoothing                         & 0.1 \tabularnewline
        param-init                              & 0 \tabularnewline
        share-embeddings                        & True \tabularnewline
        share-decoder-embeddings                & True \tabularnewline
        generator-layer-norm                    & True \tabularnewline
        warmup-steps                            & 4000 \tabularnewline
        learning-rate                           & 1 \tabularnewline
        \midrule
        \multicolumn{2}{l}{\textbf{vMF Model}} \tabularnewline
        \midrule
        dropout                                 & 0.1 \tabularnewline
        batch-size                              & 40960 \tabularnewline
        batch-type                              & tokens \tabularnewline
        normalization                           & tokens \tabularnewline
        optim                                   & radam \tabularnewline
        adam-beta2                              & 0.9995 \tabularnewline
        decay-method                            & linear \tabularnewline
        weight-decay                            & 0.00001 \tabularnewline
        max-grad-norm                           & 5.0 \tabularnewline
        lambda-vmf                              & 0.2 \tabularnewline
        generator-function                      & continuous-linear \tabularnewline
        loss                                    & nllvmf \tabularnewline
        \midrule
        \multicolumn{2}{l}{\textbf{Softmax Model}} \tabularnewline
        \midrule
        dropout                                 & 0.2 \tabularnewline
        batch-size                              & 5120 \tabularnewline
        batch-type                              & tokens \tabularnewline
        accum-count                             & 6 \tabularnewline
        optim                                   & adam \tabularnewline
        adam-beta2                              & 0.999 \tabularnewline
        decay-method                            & noam \tabularnewline
        max-grad-norm                           & 25 \tabularnewline
    \bottomrule
    \end{tabular}

\stepcounter{table}
\noindent
Table \thetable: Training hyperparameter configuration for our main softmax-based model and our main vMF-based model.

\section{Multilingual Basesystem Scores}
\label{appendix:basesystem_scores}
We evaluate all our systems using SacreBLEU \citep{post-2018-call}.
Following the best practice we list the SacreBLEU signature for our BLEU and ChrF++ evalations as follows:
\texttt{nrefs:1|case:mixed|eff:no|tok:13a|
$\hookrightarrow$smooth:exp|version:2.4.0}
\texttt{nrefs:1|case:mixed|eff:yes|nc:6|nw:2|
$\hookrightarrow$space:no|version:2.4.0}

\vspace{5px}
Table~\ref{tab:basemodel_comparison} details the ChrF++ scores of our different multilingual models averaged over their 20 supervised translation directions.
Table~\ref{tab:frozen_softmax} and \ref{tab:bpe_baseline} detail the individual scores.

\begin{table}[h!tpb]
    \centering
    \begin{tabular}[]{c|rr}
        \toprule
        model        &ChrF++& $\Delta$ \tabularnewline
        \midrule
        BPE baseline & 55.3 &          \tabularnewline
        \hline
        softmax CLWE & 55.1 & -0.16    \tabularnewline
        vMF CLWE     & 51.5 & -3.79    \tabularnewline
        \bottomrule
    \end{tabular}
    \caption{
        Comparison of average ChrF++ scores on the 20 supervised directions for the different variants of the multilingual base system.
    }
    \label{tab:basemodel_comparison}
\end{table}

\begin{table*}[htpb]
    \begin{minipage}{.5\linewidth}
    \centering
    \begin{tabular}[]{c|*{5}{r}}
        \toprule
        $\lang$ & \texttt{en} & \texttt{de} & \texttt{es} & \texttt{fr} & \texttt{it} \tabularnewline
        \hline
        \texttt{en} &      & 54.1 & 61.4 & 61.0 & 57.9 \tabularnewline
        \texttt{de} & 57.5 &      & 50.3 & 52.1 & 48.7 \tabularnewline
        \texttt{es} & 62.6 & 49.0 &      & 56.4 & 54.2 \tabularnewline
        \texttt{fr} & 61.1 & 49.2 & 54.9 &      & 53.5 \tabularnewline
        \texttt{it} & 59.7 & 48.4 & 54.8 & 55.7 &      \tabularnewline
        \bottomrule
    \end{tabular}
    \caption*{ ChrF++ }

    \end{minipage}%
    \begin{minipage}{.5\linewidth}
    \centering

    \begin{tabular}[]{c|*{5}{r}}
        \toprule
        $\lang$ & \texttt{en} & \texttt{de} & \texttt{es} & \texttt{fr} & \texttt{it} \tabularnewline
        \hline
        \texttt{en} & &30.0&39.3&44.0&36.5 \tabularnewline
        \texttt{de} &38.8& &26.5&33.6&26.1 \tabularnewline
        \texttt{es} &44.5&24.3& &38.4&32.0 \tabularnewline
        \texttt{fr} &43.1&24.8&31.5& &31.7 \tabularnewline
        \texttt{it} &41.2&23.9&31.6&37.6&  \tabularnewline
        \bottomrule
    \end{tabular}
    \caption*{ BLEU }

    \end{minipage}%

    \caption{
        ChrF++/BLEU scores of our main softmax-based model on the TED test split \citep{qi-etal-2018-pre}.
        The reported languages present the supervised training directions.
        Rows represent the source language, while columns represent the target language.
    }
    \label{tab:frozen_softmax}
\end{table*}

\begin{table*}[htpb]
    \begin{minipage}{.5\linewidth}
    \centering
    \begin{tabular}[]{c|*{5}{r}}
        \toprule
        $\lang$ & \texttt{en} & \texttt{de} & \texttt{es} & \texttt{fr} & \texttt{it} \tabularnewline
        \hline
        \texttt{en} &     &54.0&61.5&61.2&58.1 \tabularnewline
        \texttt{de} & 57.9&    &50.5&52.6&49.0 \tabularnewline
        \texttt{es} & 62.7&48.8&    &56.6&54.3 \tabularnewline
        \texttt{fr} & 61.4&49.4&55  &    &53.7 \tabularnewline
        \texttt{it} & 59.8&48.4&54.8&55.9&     \tabularnewline
        \bottomrule
    \end{tabular}
    \caption*{ ChrF++ }

    \end{minipage}%
    \begin{minipage}{.5\linewidth}
    \centering

    \begin{tabular}[]{c|*{5}{r}}
        \toprule
        $\lang$ & \texttt{en} & \texttt{de} & \texttt{es} & \texttt{fr} & \texttt{it} \tabularnewline
        \hline
        \texttt{en} &     &29.6&39.1&43.9&36.4 \tabularnewline
        \texttt{de} & 39.0&    &26.7&33.8&26.1 \tabularnewline
        \texttt{es} & 44.4&24.0&    &38.4&32.1 \tabularnewline
        \texttt{fr} & 43.3&24.9&31.5&    &31.6 \tabularnewline
        \texttt{it} & 41.3&23.9&31.5&37.6&     \tabularnewline
        \bottomrule
    \end{tabular}
    \caption*{ BLEU }

    \end{minipage}%

    \caption{
        ChrF++/BLEU scores of our BPE-based baseline model on the TED test split \citep{qi-etal-2018-pre}.
        The reported languages present the supervised training directions.
        Rows represent the source language, while columns represent the target language.
    }
    \label{tab:bpe_baseline}
\end{table*}

\section{Supplementary Zero-Shot Translation Scores}
\label{appendix:supplementary_blind_translation_scores}
Table~\ref{tab:blind_translation_bleu} details the BLEU scores for the Table~\ref{tab:blind_translation} evaluations.
Table~\ref{tab:blind_translation_full} details additional zero-shot translation scores, for languages and translation directions not listed in Table~\ref{tab:blind_translation}.
The scores are produced with our main Romance-Germanic model, so the seen languages are English, German, Spanish, French and Italian and model dimension $\dmodel = 300$.
\begin{table*}[!htpb]
    \centering
    \input{tables/blind-translation-bleu.tex}
\end{table*}
\begin{table*}[hbt!]
    \begin{minipage}{.5\linewidth}
    \centering
    \begin{tabular}[]{c|*{5}{r}}
        \toprule
        $\lang$ & \texttt{en} & \texttt{de} & \texttt{es} & \texttt{fr} & \texttt{it} \tabularnewline
        \hline
        pt & 61.0&48.4&57.1&56.2&53.6 \tabularnewline
        ru & 42.1&38.0&39.8&41.5&38.5 \tabularnewline
        pl & 40.5&36.1&38.2&39.9&36.8 \tabularnewline
        uk & 41.3&36.5&38.9&40.1&37.6 \tabularnewline
        zh & 33.3&29.0&31.2&32.1&29.7 \tabularnewline
        vi & 34.6&30.4&32.5&32.7&31.1 \tabularnewline
        ja & 28.3&26.0&27.2&27.8&26.2 \tabularnewline
        ko & 25.0&23.2&24.4&24.5&23.0 \tabularnewline
        ar & 38.3&32.1&35.8&36.3&34.0 \tabularnewline
        tr & 31.9&28.9&31.1&31.0&29.8 \tabularnewline
        nl & 51.3&45.5&46.4&48.3&44.5 \tabularnewline
        bg & 48.9&42.6&46.6&47.3&44.1 \tabularnewline
        cs & 45.2&40.0&42.2&44.2&40.9 \tabularnewline
        ro & 49.0&42.1&46.8&48.1&44.8 \tabularnewline
        \bottomrule
    \end{tabular}
    \caption*{ ChrF++ }

    \end{minipage}%
    \begin{minipage}{.5\linewidth}
    \centering

    \begin{tabular}[]{c|*{5}{r}}
        \toprule
        $\lang$ & \texttt{en} & \texttt{de} & \texttt{es} & \texttt{fr} & \texttt{it} \tabularnewline
        \hline
        pt & 42.4&23.2&34.0&38.1&31.1 \tabularnewline
        ru & 20.4&13.3&14.9&21.6&15.0 \tabularnewline
        pl & 20.0&12.1&14.4&20.5&14.0 \tabularnewline
        uk & 20.0&11.7&14.1&20.0&14.2 \tabularnewline
        zh &  9.9& 4.1& 5.3&10.4& 5.6 \tabularnewline
        vi & 12.4& 5.9& 8.0&12.1& 7.9 \tabularnewline
        ja &  4.4& 1.3& 2.2& 5.7& 2.5 \tabularnewline
        ko &  5.4& 1.6& 2.6& 5.8& 2.7 \tabularnewline
        ar & 17.0& 7.7&10.9&16.2&10.5 \tabularnewline
        tr &  9.2& 5.0& 6.9&10.4& 7.1 \tabularnewline
        nl & 31.0&20.4&22.7&29.6&21.8 \tabularnewline
        bg & 28.0&17.2&21.6&27.8&20.3 \tabularnewline
        cs & 23.1&15.1&17.5&24.6&17.1 \tabularnewline
        ro & 28.1&16.9&22.3&28.6&21.0 \tabularnewline
        \bottomrule
    \end{tabular}
    \caption*{ BLEU }

    \end{minipage}%

    \caption{
        ChrF++/BLEU scores of our main softmax-based model on the TED test split \citep{qi-etal-2018-pre}.
        The reported languages present translations from unseen source languages.
        Rows represent the source language, while columns represent the target language.
        Note that the translations are produced with a beam size of 1 and thus slightly differ from the scores reported in Table~\ref{tab:blind_translation}.
    }
    \label{tab:blind_translation_full}
\end{table*}

\section{Unsupervised MT BLEU Scores}
\label{appendix:unsupervised_mt_bleu_scores}
Table~\ref{tab:newlang_results_bleu} details the BLEU scores for the Table~\ref{tab:newlang_results} evaluations.
\begin{table*}[!htpb]
    \centering
    \input{tables/backtranslation-bleu.tex}
\end{table*}

\section{Effect of Domain}
\label{sec:effect_of_domain}
To study the effect of domain and amount of data we perform iterative back-translation training on Wikipedia and news domain.
For domain specific test data we use Flores, to evaluate the Wikipedia domain, and wmt16 test data for evaluation of the news domain.
We detail the results in Table~\ref{tab:out_of_domain_back_translation}.
When trained on Wikipedia we observe a slight increase in ChrF++ scores on English-Portuguese Wikipedia domain, but a large drop on the original TED domain.
In all other cases we observe drops in performance on all considered domains.
\begin{table*}[!pb]
    \centering
    \begin{tabular}{clrrrr}
        \toprule
                                            & train on   & \#sentences  & TED  & Flores & wmt16 \tabularnewline
        \midrule
        \multirow{2}{*}{en$\rightarrow$tr}  & TED        &  182k & 41.1 & 38.7 &      \tabularnewline
                                            & Wiki-40B   & 1947k & 32.8 & 37.8 &      \tabularnewline
        \midrule
        \multirow{2}{*}{en$\rightarrow$pt}  & TED        &   51k & 57.3 & 50.5 &      \tabularnewline
                                            & Wiki-40B   & 6094k & 51.6 & 51.8 &      \tabularnewline
        \midrule
        \multirow{2}{*}{en$\rightarrow$ro}  & TED        &  180k & 49.1 & 48.0 & 43.2 \tabularnewline
                                            & NewsCrawl  & 2281k & 44.2 & 44.9 & 43.9 \tabularnewline
        \bottomrule
    \end{tabular}
    \caption{A comparison of average test scores for different variations of the autoencoder methods (top), as well as the back-translation method (bottom).}
    \label{tab:out_of_domain_back_translation}
\end{table*}

\section{Estimate of GPU Hours}
\label{sec:estimate_of_gpu_hours}
We conduct our preliminary experiments on an \texttt{Nvidia GTX 1080Ti} GPU.
The main body of our experiments if conducted on a cluster of \texttt{Nvidia Titan RTX} GPUs.
Our largest softmax-based full-word vocabulary model (with a count of 53M trainable parameters, 317M non-trainable) is trained on an \texttt{Nvidia A6000} GPU.
The GPU hours roughly sum up to 1039 GPU hours, 197h out of these on an \texttt{Nvidia A6000} GPU.

%% file: tables/blind-translation-bleu.tex
\begin{adjustwidth}{-0.25cm}{}
\setlength{\tabcolsep}{4pt}
\begin{tabular}{l|rc|rrrrrrrrrrr}
    \toprule
    Trained on         &\small $|\baselangs|$ &\small $\dmodel$ & de-en & es-en & ru-en & vi-en & pt-en & nl-en & bg-en & ar-en & tr-en & ko-en \tabularnewline
    \toprule
    \multicolumn{13}{l}{\textbf{Softmax Models}} \tabularnewline
    \toprule
    \tikzmark{bleu_bpe_from}%
    BPE baseline                           & 5 & 300 & \grcll 39.0 & \grcll 44.4 &  2.1 &  2.0 & 12.9 &  5.6 &  2.0 &  2.7 &  2.9 &  3.5 \tabularnewline
    \tikzmark{bleu_bpe_to}%
    \citep{garcia-etal-2021-towards}       & 5 & 300 & \grcll & \grcll & \grcll 27.4 & \grcll 29.7 & \grcll 44.0  & \grcll 38.6 & \grcll 40.9 & \grcll 32.0 & \grcll 27.7 & \grcll 21.4 \tabularnewline
    \midrule
    \tikzmark{bleu_softmax_from}%
    \small en,de,es,fr,it                  & 5 & 300 & \grcll 38.8  & \grcll 44.5  & 20.7  & 13.0  & 42.6  & 31.5  & 29.0  & 17.6  &  9.5  &  5.6   \tabularnewline
    \small en,de,es,fr,it,ru,uk,pl,vi,zh   &10 & 576 & \grcll 37.1  & \grcll 43.1  & \grcll 28.0  & \grcll 28.7  & 41.6  & 31.2  & 33.0  & 21.5  & 11.9  & 7.7 \tabularnewline
    \midrule
    \tikzmark{bleu_softmax_to}%
    supervised fine-tune                   & 5 & 300 & \grcll & \grcll & \grcll 28.3 & \grcll 27.4 & \grcll 45.2 & \grcll 37.8 & \grcll 39.6 & \grcll 33.5 & \grcll 27.7 & \grcll 19.3 \tabularnewline
    \toprule
    \multicolumn{13}{l}{\textbf{vMF Models} \citep{kumar2018vmf}} \tabularnewline
    \midrule
    \texttt{es}$\rightarrow$\texttt{en}    & 1 & 300 & 11.4  & \grcll 42.4  & 12.8  &  5.7  & 34.2  & 16.6  & 12.0  &  8.8  &  5.5  &  3.4 \tabularnewline
    \texttt{es}$\leftrightarrow$\texttt{en}& 2 & 300 & 16.8  & \grcll 41.8  & 15.6  &  7.8  & 35.3  & 20.4  & 18.0  & 11.9  &  7.2  &  4.2 \tabularnewline
    \midrule
    \small en,de,fr,it                     & 4 & 300 & \grcll 36.0  & 35.8  & 18.3  & 10.0  & 36.9  & 28.0  & 24.3  & 13.3  &  7.8  &  4.5  & \tabularnewline
    \tikzmark{bleu_vmf_from}%
    \small en,de,es,fr,it                  & 5 & 300 & \grcll 36.0  & \grcll 41.7  & 18.9  & 10.2  & 39.6  & 28.3  & 24.0  & 14.5  &  8.3  &  4.5 \tabularnewline
    \small en,de,es,fr,it,ru,uk,pl         & 8 & 300 & \grcll 34.5  & \grcll 40.0  & \grcll 25.9  & 12.6  & 37.8  & 28.6  & 30.5  & 16.3  & 10.1  &  6.3 \tabularnewline
    \small en,de,es,fr,it,ru,uk,pl,vi,zh   &10 & 300 & \grcll 33.6  & \grcll 39.2  & \grcll 25.3  & \grcll 26.4  & 36.5  & 28.0  & 29.7  & 16.5  & 10.1  &  6.6 \tabularnewline
    \midrule
    \tikzmark{bleu_vmf_to}%
    supervised fine-tune                   & 5 & 300 & \grcll 36.1  & \grcll 42.0  & \grcll 25.7  & \grcll 25.3  & \grcll 60.4  & \grcll 35.4  & \grcll 38.1  & \grcll 29.0  & \grcll & \grcll 15.0 \tabularnewline
    \bottomrule
\end{tabular}
\end{adjustwidth}
\caption{
    ChrF++ scores for decoding from an unseen language on the TED data test split \citep{qi-etal-2018-pre}.
    Grey colored cells indicate supervised training on the source language, white cells indicate no prior exposure to the source language.
    $\dmodel$ denotes the Transformer model dimension and arrows indicate what models are used as basis for fine-tuning.
    Note, that every column in the three fully supervised rows represents a separate bilingual fine-tune. 
    For BLEU scores and translation directions other than English see Appendix~\ref{appendix:supplementary_blind_translation_scores}.
}
\label{tab:blind_translation_bleu}

\begin{tikzpicture}[overlay,remember picture]
\draw[->] ($(pic cs:bleu_bpe_from)+(-4pt,.5ex)$) to ($(pic cs:bleu_bpe_from)+(-10pt,.5ex)$) to ($(pic cs:bleu_bpe_to)+(-10pt,+0.5ex)$) to ($(pic cs:bleu_bpe_to)+(-4pt,+0.5ex)$);
\draw[->] ($(pic cs:bleu_softmax_from)+(-4pt,.5ex)$) to ($(pic cs:bleu_softmax_from)+(-10pt,.5ex)$) to ($(pic cs:bleu_softmax_to)+(-10pt,+0.5ex)$) to ($(pic cs:bleu_softmax_to)+(-4pt,+0.5ex)$);
\draw[->] ($(pic cs:bleu_vmf_from)+(-4pt,.5ex)$) to ($(pic cs:bleu_vmf_from)+(-10pt,.5ex)$) to ($(pic cs:bleu_vmf_to)+(-10pt,+0.5ex)$) to ($(pic cs:bleu_vmf_to)+(-4pt,+0.5ex)$);
\end{tikzpicture}

%% file: tables/backtranslation-bleu.tex
\setlength{\tabcolsep}{2pt}
\begin{tabular}{l rcl r rcl r rcl r rcl}
    \toprule
    \multirow{2}{*}{Model}             & \multicolumn{3}{c}{en$\rightarrow$pt} & & \multicolumn{3}{c}{en$\rightarrow$ru} & & \multicolumn{3}{c}{en$\rightarrow$tr} & & \multicolumn{3}{c}{en$\rightarrow$ko} \tabularnewline
    \cline{2-4}\cline{6-8}\cline{10-12}\cline{14-16}
                                       & \small TED & \small Flores & \small Tatoeba & & \small TED & \small Flores & \small Tatoeba & & \small TED & \small Flores & \small Tatoeba & & \small TED & \small Flores & \small Tatoeba \tabularnewline
    \midrule
    \citep{song2019mass}  &       &       &       & &        &       &       & &   4.0  &  4.8  &  1.3  & &        &       &      \tabularnewline
    \citep{artetxe-etal-2019-effective}
                          &       &       &       & &        &       &       & &   0.4  &  0.2  &  0.3  & &        &       &      \tabularnewline
    \midrule
    Iteration 1           & 36.8  & 27.4  & 33.4  & & 17.4  & 14.1  & 25.2  & &  8.1  &  5.4  &  8.2  & &  8.0  & 7.0  &  7.9  \tabularnewline
    Iteration 3           & 37.4  & 28.0  & 33.7  & & 18.5  & 15.7  & 27.5  & & 11.0  &  6.9  & 13.8  & & 11.4  & 8.6  &  9.8  \tabularnewline
    Iteration 5           & 37.3  & 27.9  & 33.7  & & 18.7  & 16.4  & 29.9  & & 12.5  &  7.6  & 16.5  & & 12.5  & 9.4  & 10.8  \tabularnewline
    \midrule
    Supervised            & 36.4  & 26.8  & 32.2  & & 20.1  & 16.5  & 29.6  & & 15.7  & 10.2  & 21.1  & & 14.0  & 11.2  & 13.8 \tabularnewline
    \toprule

    \multirow{2}{*}{Model}             & \multicolumn{3}{c}{pt$\rightarrow$en} & & \multicolumn{3}{c}{ru$\rightarrow$en} & & \multicolumn{3}{c}{tr$\rightarrow$en} & & \multicolumn{3}{c}{ko$\rightarrow$en} \tabularnewline
    \cline{2-4}\cline{6-8}\cline{10-12}\cline{14-16}
                                       & \small TED & \small Flores & \small Tatoeba & & \small TED & \small Flores & \small Tatoeba & & \small TED & \small Flores & \small Tatoeba & & \small TED & \small Flores & \small Tatoeba \tabularnewline
    \midrule
    \citep{song2019mass}  &       &       &       & &        &       &       & &   4.1  &  5.9  &  2.9  & &        &       &      \tabularnewline
    \citep{artetxe-etal-2019-effective}
                          &       &       &       & &        &       &       & &   0.2  &  0.1  &  0.1  & &        &       &      \tabularnewline
    \midrule
    Zero-shot             & 42.6  & 30.6  & 38.7  & & 20.7  & 13.8  & 24.8  & &  9.5  &  5.0  &  6.4  & &  5.6  & 2.5  & 3.0  \tabularnewline
    Iteration 2           & 44.0  & 31.1  & 40.6  & & 24.9  & 17.0  & 31.8  & & 15.6  &  9.3  & 17.9  & &  9.6  & 4.1  & 7.2  \tabularnewline
    Iteration 4           & 43.6  & 30.8  & 39.9  & & 25.9  & 18.5  & 35.6  & & 19.0  & 10.8  & 24.5  & & 10.5  & 4.4  & 8.0  \tabularnewline
    Iteration 6           & 43.5  & 30.8  & 39.5  & & 26.0  & 18.7  & 36.2  & & 21.1  & 12.1  & 28.6  & & 10.8  & 4.6  & 8.4  \tabularnewline
    \midrule
    Supervised            & 45.2  & 31.0  & 41.3  & & 28.3  & 19.2  & 39.4  & & 27.7  & 17.5  & 35.7  & & 19.3  & 9.6  & 21.8 \tabularnewline
    \bottomrule
    \# of Sentences        & \multicolumn{3}{c}{52k} & & \multicolumn{3}{c}{208k} & & \multicolumn{3}{c}{182k} & & \multicolumn{3}{c}{206k} \tabularnewline
    \bottomrule
\end{tabular}

\caption{
    BLEU scores after each iteration of back-translation (see Table~\ref{tab:newlang_results}).
    In each iteration we use our Romance-Germanic multilingual model to zero-shot translate from TED domain to create synthetic parallel data.
}
\label{tab:newlang_results_bleu}